**Title:** Lung airway geometry as an early predictor of autism: A preliminary machine learning-based study.

**Authors:** Asef Islam[1,2], Anthony Ronco[3], Stephen M. Becker[4], Jeremiah Blackburn[4], Johannes C. Schittny[5,6], Kyoungmi Kim[7], Rebecca Stein-Wexler[3], Anthony S. Wexler[4,8,9,10]

**Affiliations:**

1. Department of Computer Science, Stanford University, Stanford, CA, USA
2. Department of Biomedical Informatics, Stanford University School of Medicine, Stanford, CA, USA
3. Department of Radiology, University of California, Davis, CA, USA
4. Department of Mechanical and Aerospace Engineering, University of California, Davis, CA, USA
5. Institute of Anatomy, University of Bern, Bern, Switzerland
6. Center for Health and the Environment, University of California, Davis, CA, USA
7. Department of Public Health Science, University of California, Davis, CA, USA
8. Department of Civil and Environmental Engineering, University of California, Davis, CA, USA
9. Department of Land, Air and Water Resources, University of California, Davis, CA, USA
10. Air Quality Research Center, University of California, Davis, CA, USA

**Abstract**

The goal of this study is to assess the feasibility of airway geometry as a biomarker for ASD. Chest CT images of children with a documented diagnosis of ASD as well as healthy controls were identified retrospectively. 54 scans were obtained for analysis, including 31 ASD cases and 23 age and sex-matched controls. A feature selection and classification procedure using principal component analysis (PCA) and support vector machine (SVM) achieved a peak cross validation accuracy of nearly 89% using a feature set of 8 airway branching angles. Sensitivity was 94%, but specificity was only 78%. The results suggest a measurable difference in airway branchpoint angles between children with ASD and the control population.

**Introduction**

Autism spectrum disorder (ASD) refers to a group of heterogenous neurodevelopmental disorders characterized by impairments in communication and social interaction as well as restricted, repetitive, and stereotypic behavior patterns (1. American Psychiatric Association, 2013). ASD affects more than 1% of children born in the US and is estimated to cost the US $61 billion per year (Buescher, 2014). Furthermore, the prevalence of ASD appears to be increasing (Prevention, 2014)[3].

Early diagnosis and treatment of autism spectrum disorder (ASD) continues to represent a significant clinical and public health challenge. Numerous studies have proved that early identification and intervention results in improved outcomes in children with ASD (4. National Research Council, 2001; Frazier, 2021). Although the prevalence of ASD appears to be increasing, the average age of diagnosis remains static at 3 years of age (Center for Disease Control and Prevention, 2007; Robins, 2008). This is likely due to a combination of factors, one of which is the methods of screening and diagnosis. The current standards for screening and diagnosing ASD involve social and behavioral observations in a structured setting, typically after the age of two years (Zwaigenbaum, 2009). While tools such as the Infant Toddler Checklist (ITC) and Modified Checklist for Autism in Toddlers-Revised (M-CHAT-R) have been well-validated, they rely on collecting data that does not manifest until a child is old enough to exhibit the appropriate behavioral traits (Wetherby, 2002; Robins, 2014). ITC is useful for assessing developmental deficits in children ages 9-24 months, but it may not distinguish basic communication delays from ASD (Wetherby, 2008). The M-CHAT-R, a validated screening tool for toddlers between 16 and 30 months of age, requires a follow up questionnaire for positive screens, delaying diagnosis (Robins, 2014). The resultant delay in intervention may have devastating consequences and is a missed opportunity to improve outcomes for these patients, likely at a high cost on both a personal and societal level given the high prevalence of disease.

Multiple co-morbidities have been linked with (ASD), including schizophrenia, epilepsy, sleep disorders, and a range of gastrointestinal disturbances (Mannion, 2013). Chronic neuroinflammatory and immune dysfunction have also been described in association with ASD (Matta, 2019; Bjorklund, 2016). Given the pervasive nature of the disease and its presence during embryonic development, biomarker screening in conjunction with behavioral diagnostic methods is an attractive solution to the problem of early detection. Biomarkers associated with ASD are an active area of research. Associations have been made between ASD and head shape, gut microbiome, and serum micro-RNA profile (Mannion, 2013). Many physiologic markers not necessarily directly connected to behavior— such as head circumference growth and finger length ratio— have been proposed as early indicators of autism (Elder, 2008; De Bruin, 2006). There is a significant genetic component to ASD, with most data supporting a polygenic epistatic model (Persico, 2013; Cristino, 2014). Due to the heterogeneous nature of this disorder, classical genetic studies have not been successful in identifying candidate genes for ASD. In recent years, epigenetic mechanisms, which control heritable changes in gene expression without changing the DNA sequence, have been implicated as a potential contributor to the pathogenesis of ASD (Miyake, 2012). Studies have shown the epigenetic role of microRNA could offer insights into the pathogenesis of autism (Vasu, 2014; Hicks, 2016). However, these biomarkers have insufficient sensitivity and specificity to successfully predict ASD.

In a recent study, Stewart and Klar proposed lung airway geometry as a physiological biomarker of autism (Stewart, 2013). In brief, they found that branching anomalies identified during bronchoscopies, termed "doublets", strongly correlate with a diagnosis of ASD. Doublets are defined as symmetric double branching of bronchi in generation 3 and beyond. Stewart and Klar performed a retrospective study of bronchoscopies of 49 children with and 410 without a diagnosis of ASD. Remarkably, all 49 autistic children had doublets while none of the 410 neurotypic children did. This is important because airway geometry—including doublets—is formed in utero and the relative dimensions do not change with subsequent lung development (Burri, 1984). Therefore, it may be possible to identify an abnormal branching pattern in the early postnatal period in at-risk children, allowing early diagnosis of ASD. Respiratory co-morbidities with ASD have been found by others. For instance, Biot's and Cheyne–Stokes respiration and other respiratory dysrhythmias have been associated with ASD (McCormick, 2018). And a number of studies have shown associations between Respiratory Sinus Arrhythmias and ASD (Schittny, 2017), including at least one study showing predictive ability (Koos, 2014).

Pertinent to this question of pulmonary biomarkers of ASD is previous work conducted by our group in developing models and algorithms for automated extraction of airway geometry parameters from CT scans. Software has been developed to perform a segmentation of airways from a CT scan, on which a flexible bifurcation model can be iteratively fit to each branching point in the airway tree (Lee, 2008b; Lee, 2011). The resulting parameters are reported from this best-fit optimization, including the radii, diameters, and lengths of parent and daughter branches and branching angles at each bifurcation. Previously this method has been used to reveal structural changes in developing lung geometry due to insults to pulmonary health such as particle inhalation in rats and other model organisms (Lee, 2010, 2011abc; Islam, 2017). The implication of applying this to the present ASD setting is that presumably these doublets or other intrinsic morphological indicators of ASD in airway structure may be evident in the measurements identified by this software. Advancements in high-resolution in-vivo imaging techniques combined with the development of powerful statistical learning methods may lead to a highly reliable and automated diagnostic tool for ASD from lung scans.

The study presented here provides quantitative, statistical analysis of airway geometric parameters predictive of autism, derived from chest computed tomography (CT) scans. To our knowledge it is the only other study specifically evaluating anatomic airway anomalies in the setting of ASD, though other research studies have assessed the molecular and pathophysiologic implications of the findings of Stewart and Klar (Brennan, 2013; Ning, 2019, Jónsdóttir, 2017).

**Materials and Methods**

**Ethics statement**

This retrospective study was approved by the Institutional Review Board for the Protection of Human Subjects (IRB) of the University of California Davis Health System (UCDHS). Written consent was not obtained, since the retrospective study did not affect the care of the included individuals. All patients' records/information were anonymized and de-identified prior to analysis.

**Patient selection**

The inclusion criteria in this study are male or female patients ages 1-30 with or without a documented history of ASD who underwent chest CT with ~1 mm resolution images. Patients with history of congenital cardiac or pulmonary anomalies, sub-optimal CT images, or missing or incomplete medical records were excluded. 31 subjects with ASD (age = 15.0 ± 7.4 years [mean ± SD]; range = 4-30 years; 21 males and 10 females) and 23 age- and sex- matched control subjects (age = 16.2 ± 3.3 years; range = 9-20 years; 12 males and 11 females) were included. There were no significant differences in the age (p= 0.43) or sex (p=0.25) distributions between the control and ASD groups. The electronic medical record was searched for terms including "autism," "ASD," and "autism spectrum disorder" along with concomitance of a chest CT to identified verified case studies. The control CTs were identified from the clinical pediatric radiology work lists. Table 1 summarizes patient statistics for this study.

**3D limited flood fill of airways using custom software**

In our prior work on the geometry of rat airways (Lee, 2008ab, 2010, 2011abc), we removed the lungs from the rats, filled them with silicone, eroded the tissue, and performed CT scans of the lung casts. This was advantageous for rat imaging because there was no motion artifact and we could use long exposure times to image at high resolution. To obtain geometric parameters for human conducting airways, we developed software that performed a 3D limited flood fill (3DLFF) of the airways, imitating in software the silicone casting procedure used for imaging rat airways. 3D flood filling is a standard algorithm used to fill shapes in 3 dimensions (C implementation; Didactic; Zhong, 2016; QuickFill; Sample implementation; Flash Flood). But our 3D flood fill had to be more sophisticated than standard flood filling in a few ways: (i) The airways from the trachea down had to be filled uniformly and completely, similar to silicone filling the rat airways in our prior studies. To accomplish this, a standard 3D flood fill algorithm was used, but the position of the front was tracked as it propagated through the distal conducting airways. New voxels were always filled uniformly across the front. (ii) The flood-fill algorithm limited the number of voxels to fill so that the filling would complete without "spilling" into the parenchyma. (iii) In order to limit radiation exposure, the resolution of human chest CT scans is usually about 1 mm between planes and 0.5 mm within plane. As a result, as airways turn and become parallel to the scan plane, apparent holes in the bronchiole wall can develop that enable the fill process to spread into the surrounding lung parenchyma. The 3DLFF code has two features to deal with this: First, voxels will not be filled if they are detected to be in a hole of a user-selected size. Smaller holes are not filled, eliminating some leaks. Second, if leaks are found, the code has a mechanism for the user to manually plug holes.

The 3D flood fill code was used as follows: First, the code was used to fill the parenchyma, plugging many of the airway holes from the outside. Then the code was used to fill the airway lumen starting at the trachea. Figure 1 shows selected slices from one subject before and after this filling process.

**Airway geometric parameters**

We have developed software that extracts lung geometry from CT images of conducting airways. Each conducting airway bifurcation is modeled as a straight section, followed by a smooth transition to two offspring airways that are modeled as sections of toroids (see Figure 2). Each bifurcation is characterized by 15 parameters: length and diameter of the parent (2), subtended angle, radius of curvature, and diameter of each of the two offspring (2x3=6); 3 positional parameters and 3 orientational parameters; and the carina radius (Lee, 2008a). Vectors are then drawn perpendicular to the surface of the model that extend inwards or outwards until they intersect the image of the airway in the CT scan. A minimization algorithm adjusts 14 of the 15 parameters (we fix the carina radius because the image resolution is insufficient to resolve it) until the distance between the model and the image is as small as possible (Lee, 2011). This process is carried out iteratively down the airway tree, bifurcation by bifurcation, until the resolution of the image is insufficient to obtain accurate geometric information. This typically happens when the airway diameter is smaller than 6 voxels, which is 3 mm in human CTs that have a 0.5 mm in-plane resolution. The result is 14 geometric parameters for the bifurcations that are resolved. In our preliminary work, we were typically able to resolve 13 to 15 bifurcations for each subject for a total of about 196 geometric parameters.

**Machine Learning Approach**

With high-dimensional classification tasks such as this, machine learning algorithms may prove useful in learning a decision rule to differentiate between positive and negative examples in the data. Support Vector Machines (SVMs) are a popular supervised learning method that in the healthcare domain alone have been used in predicting ailments such as diabetes, cancer, neurodegenerative diseases such as Alzheimer's and orthopedic conditions such as osteoarthritis (Razzaghi, 2016; Yu, 2010; Battineni, 2019; Charon, 2021). While more powerful methods exist, SVMs benefit from being simple linear models with high interpretability. In addition, more advanced methods such as deep learning generally require significantly more training data than we currently have. Essentially, SVM operates by seeking the hyperplane that separates the two classes with maximum margin among labelled data points in $m$-dimensional space, where $m$ is the number of features given as input. We sought to develop a feature selection/extraction process and SVM tuning procedure to identify a subset of geometric parameters that were most informative for determining ASD condition and test performance of a linear SVM in learning this decision boundary.

We first restricted the feature space to just angles in the first four branching generations based on prior clinical background on the locations of relevant biomarkers for ASD (Stewart, 2013). Thus, the total number of possible parameters is 26 (13 bifurcations in the first 4 generations * 2 branching angles per bifurcation). These angles were first normalized (for each angle, we subtracted the mean of all measurements and then divided by the standard deviation of all measurements). While it is possible to simply input all these features into training the SVM model, often performance is increased by first conducting feature engineering and filtering down to a smaller subset of features that preserves the most information. This is due to what is often referred to as the "curse of dimensionality", which is a situation in which having an excessively high-dimensional problem (too many features relative to the number of examples) can actually hurt performance because features contain redundant information and add noise to the model. Thus, it is advantageous to select a smaller subset of features that still preserves as much unique information as is relevant for the classification task.

A common feature selection algorithm is known as principal component analysis (PCA). PCA is a generalized dimensionality reduction algorithm in which singular value decomposition (SVD) is performed on the data matrix to decompose it into the eigenvectors of the corresponding covariance matrix. This means that the data matrix becomes decomposed into a series of orthogonal eigenvectors ordered by the amount of variance that they represent of the original data. This can thus be used to obtain lower-dimensional reductions of the data with the maximum preservation of variance in the data. For example, by performing PCA on an *n x m* data matrix and selecting the first *p (p < m)* principal components of the resulting decomposition, we reduce the original *m*-dimensional data to the *p*-dimensional representation that preserves as much as the original variance (i.e. discriminatory information) from the data as possible.

We combine the PCA feature selection step with the SVM training step into a pipeline to tune the classification procedure as follows: We first loop through *k* from 1 to *m*, where *m* is the total number of features in our selection space (in this case 26 for the total number of angles being considered). For each value of *k*, we use PCA to reduce the data to the *k* dimensional representation of maximum variance preservation by selecting the first *k* principal components. Then, within each iteration of this loop we perform a leave one out cross validation (LOOCV) experiment to determine the classification accuracy of a linear SVM on the data. This means that we loop through each example in our data, and for each one we leave it out and train an SVM classifier on all of the other examples, and then use the left-out example as a test subject to determine whether our classifier predicts its label correctly or not. Then, the final accuracy metric is based on the number of examples that were classified correctly when used as a left-out test subject. We have one accuracy score for each iteration of the outer loop, i.e. each number of possible principal components that we tried from 1 to *m,* which we can use to select the dimensionality that performed best.

This procedure was taken further by adding an additional feature selection step in the beginning. Given that we know it is particular bifurcation angles that are likely most significant for displaying ASD-like anomalies, as well as the fact that it is often wise in machine learning generally to heuristically pre-select features based on domain-specific knowledge, we aimed to filter down the features to a smaller set of angles that would be used in the test for ASD. Furthermore, in a clinical setting, a test requiring only a few angles to be measured rather than dozens would be beneficial. However, a greater number of angles may provide more information, and thus a tradeoff is needed to find the optimal number of features that provides as much information as needed and no greater. In this case, we took a somewhat brute force, exhaustive searching approach to determine the subset of angles that would be most useful. First, we restricted the search space further to exclude generations 1 and 2, thus now we are only selecting from generations 3 and 4 (20 angles total) since those are where the most relevant angles are located (Stewart, 2013). Then, we searched by iterating through *i* from 1 to 20, and for each *i* trying every possible subset of angles of size *i* from the 20 available angles. For each of these subsets, we use it as the feature space for the PCA-SVM pipeline described previously. Thus, the overall procedure is to loop through each possible subset size *i*, then for each possible subset of that size loop through each dimensionality *k*, and perform LOOCV to determine the highest accuracy and thus best-performing model. One complication is that the number of possible subsets to search over increases significantly as *i* increases because the number of possible subsets is equal to $\binom{20}{i}$. For example, there are 15,504 possible subsets of size 5, and 38,760 subsets of size 6. The exhaustive search was conducted up to size 8, however beyond that point it became too computationally intensive to pursue to larger sizes. Figure 3 summarizes in pseudocode the entire procedure that was performed.

One worthwhile note is that a pattern emerges that allows for a method of heuristically extrapolating to larger size subsets more efficiently. Up to size 6, the best-performing subset of size *i* is generally all of the angles from the best subset of size *i-1*, plus one of the other remaining angles. Thus, if we assume this pattern holds true, then rather than exhaustively searching all possible subsets, we simply take the best subset from the previous iteration and test all subsets created from adding one other additional angle to that subset, so we are only testing 20-*i* subsets rather than $\binom{20}{i}$. We then repeat this procedure to grow until a subset size of 20. Using this approach, however, no better performing subsets were found. Of course, this does not guarantee that there are other, better-performing large subsets that are missed by this method.

In order to produce a prediction on a single example *x* of dimensionality *p* (a *p*-vector), i.e. a set of *p* angles, one would proceed as follows. First, normalize the example by subtracting the training means of each angle and dividing by the training standard deviations of each angle.

$$x' = (x - \bar{x}) \odot \frac{1}{\sigma}$$

given a *p x p* projection matrix P given by performing PCA, and a *p*-vector decision boundary *w* given by SVM, the decision rule is then

$$w \cdot Px' > 0$$

Here, multiplying *P (p x p)* by *x' (p x 1)* produces a *p*-vector, and taking the dot product of that with *w* produces a value which if greater than 0 results in a positive diagnosis and otherwise gives a negative diagnosis.

**Manual airway analysis**

Selected angles from each patient were also evaluated manually by a radiology resident with 3 years of experience in clinical imaging. The radiologist was blinded to the case/control status to remove bias. The angles of interest identified by the machine learning algorithm described above were provided, and the radiologist identified them on the chest CT using a similar workflow to the software algorithm. The trachea was identified first, then each branchpoint was determined by the sizes of the daughter airways until the appropriate bronchial branching generation was reached. The angles were measured using multiplanar reconstruction (MPR) and a modified double-oblique technique to ensure accurate measurements between parent and daughter airways (Achenbach, 2013).

We then sought to apply the decision rules derived from the machine learning approach to diagnosing cases using manually measured angles as inputs. To do this, we tried using both a subset of 3 angles, B121A1, B122A2, and B1121A2 as well as a larger subset of 5 angles comprised of the previous 3 plus B1121A1 and B1111A2. These codes are defined as follows (see Figure 4): "B" means branch, the first number is always 1 – it is the trachea. The subsequent numbers are 1 or 2, where 1 is the larger of the airways at each branch point. "A" means angle of the branch distal to the prior airway and the following number, 1 or 2, identifies which angle is being measured. Angles are measured between the direction of the parent branch and the offspring branch. The projection matrices and decision vectors associated with these subsets are:

$$P_3 \approx \begin{matrix} 0.6428 & -0.2779 & 0.7138 \\ 0.2679 & 0.9546 & 0.1303 \\ 0.7176 & -0.1075 & -0.6881 \end{matrix} \qquad w_3 = \begin{matrix} -0.1594 \\ -0.1612 \\ -0.5596 \end{matrix}$$

$$P_5 \approx \begin{matrix} 0.4850 & -0.2411 & -0.3838 & 0.7339 & -0.1442 \\ 0.1515 & -0.3063 & 0.9001 & 0.2497 & -0.1032 \\ 0.6673 & 0.0180 & 0.0489 & -0.2737 & 0.6907 \\ -0.0893 & 0.8204 & 0.1900 & 0.4748 & 0.2395 \\ 0.5372 & 0.4179 & 0.0634 & -0.3141 & -0.6588 \end{matrix} \qquad w_5 = \begin{matrix} -0.4109 \\ -0.8083 \\ -0.5690 \\ 0.6330 \\ -0.1499 \end{matrix}$$

**Results**

The result of the LOOCV procedure run on the PCA-SVM pipeline with all 26 angles is shown in Figure 5, where the number of principal components used is plotted against the resulting LOOCV accuracy. There is some fluctuation in score across the number of PCs with the optimal accuracy being around 77% with 14 principal components used. Furthermore, this optimal model had a sensitivity of 86% and a specificity of 61%. It is worth noting that the optimal performance occurs somewhere in the middle of the range of dimensionality rather than increasing with more components added, because for aforementioned reasons more dimensions may degrade performance.

When the feature space was restricted further to just the 3rd and 4th airway generations, performance improved further to an accuracy of 83% with sensitivity of 91% and specificity of 67%.

The angle subset search approach was then conducted. For each possible subset size from 1 to 8, each possible subset of angles of that size was tested in turn as the feature set. The PCA-SVM LOOCV process was repeated for each subset to identify which of the possible number of principal components would result in the highest accuracy, as was done with the full angle set. Thus, the accuracy, sensitivity and specificity of the optimal model for each subset were gathered. Table 2 summarizes this as the best-performing subset for each size. The peak overall accuracy of nearly 89% was achieved using angles B112A1, B1111A1, B111A2, B112A2, B1111A2, B1121A2, B1122A2, B1212A2. However, an accuracy of nearly 87% could be achieved using just angles B1112A1, B1211A1, B1121A2, B1212A1. Figure 6 plots peak LOOCV accuracy as a function of angle subset size.

In applying the machine learning-derived decision rules to manually measured angles, accuracy achieved was 64% using 3 angles and 54% using 5 angles. Sensitivities were 33% and 29% for 3 and 5 angles respectively, and specificities were 82% and 68% respectively.

**Discussion**

Autism spectrum disorder is a lifelong neurodevelopmental disorder with increasing prevalence. Over the last decade the autism rate has increased, currently affecting 1:68 children (Center for Disease Control and Prevention, 2008). Several studies have dispelled the notion that this rising prevalence is due to increased awareness, improved diagnosis, and changes to the diagnostic criteria (Grether, 2009; Hertz-Piccciotto, 2009; King, 2009). The current high ASD rate will have long-term costly and devastating consequences for families and society (Ganz, 2007). Numerous studies suggest that early intervention can improve outcomes for ASD children and that the earlier the intervention, the better the outcome (Reichow, 2014; Goin-Kochel, 2007; Myers, 2007; Harrington, 2014). Unfortunately, the first signs of ASD are typically deficits in communication and language, which may not manifest until year 2 of life (Fernell, 2013). Current screening methods rely on parental questionnaires that are less than 50% specific and not valid until 18-24 months of age. This approach is inefficient in identifying children with ASD and enrolling them in early intervention services (Daniels, 2014).

A number of non-behavioral diagnostic biomarkers of ASD have been proposed. Head circumference growth rates in autistic children vary significantly from neurotypic children during the first two years of life (Elder, 2008). The ratio between the lengths of the second and fourth digits also differs between autistic and neurotypic children (De Bruin, 2006). Unfortunately, the non-autistic parents and siblings of autistic children share this length ratio difference (Manning, 2001). Calcium signaling abnormalities may also have a role in both the development of autism and in disrupting normal lung development, indicating a possible link between ASD and lung development (Gargus, 2009). Hypothyroxinemia has been associated with autism and could be diagnostic (Pearce, 2013). In recent years, epigenetic mechanisms—which control heritable changes in gene expression without changing the DNA sequence—have been implicated as potential contributors to the pathogenesis of ASD (Miyake, 2012). Studies have shown the epigenetic role of serum and saliva microRNA could offer insights in the pathogenesis of autism (Vasu, 2014; Hicks, 2016). So far, these non-behavioral biomarkers have insufficient sensitivity and specificity for use as a clinical diagnostic tool.

In a recent study, lung airway anomalies were associated with autism (Stewart, 2013). While performing routine bronchoscopies on children with respiratory symptoms, Stewart and Klar noticed unusual airway geometry in autistic children. They termed these abnormal geometries "doublets," which are characterized by symmetric double branching of bronchi in generation 3 and beyond. In a blinded study of 459 children who had bronchoscopies on file, 410 were neurotypic and 49 were autistic. Remarkably, all of the autistic children had doublets while none of the neurotypic patients did. Although Stewart and Klar did not point this out, the ratio of 49 autistic children to 459 total is about 10%, indicating that autistic children are much more likely to have respiratory symptoms requiring a bronchoscopy than neurotypic children. This suggests that the doublets could also result in abnormal lung function or impaired immunologic mechanisms. Other studies support this contention. Respiratory sinus arrhythmia (RSA) is associated with both social behavior and cognitive function in children with autism (Patriquin, 2013). Respiratory dysrhythmias have been detected in children with autism (Ming, 2016). Respiratory allergies are more common in autistic children compared to the neurotypic population (Gurney, 2006). Mitochondrial respiratory chain disorder has been found in 7.2% of autistic children (Oliveira, 2005).

The machine learning method presented here demonstrated a high accuracy using a model trained with this dataset. With relatively few airway branching angles taken as input, we were able to train models that demonstrated LOOCV accuracies of up to 89%. This provides ample support to the proposition that it would be possible to develop a reliable clinical test to predict ASD from such airway branching angle measurements. However, there was a general disparity between sensitivity and specificity, with most of the models demonstrating a significantly higher sensitivity than specificity. For the model with 89% overall accuracy, for example, the sensitivity was 94% while specificity was only 78%. In the clinical setting, this would translate to a test that is effective at detecting ASD when it is present, but also prone to false positives. Of course, ideally both of these statistics should be high, and while overall accuracy was used as the primary metric of performance there were some models observed that had higher specificities and more parity between sensitivity and specificity but at slightly lower accuracy than alternative models. In designing a final test, it may be possible to select a model that has better specificity, or to add an additional confirmatory step or design a multi-layered test to reduce the false positive rate.

Of course, there are limitations to this approach that should be considered and further work to do before a reliable test can be accepted in a clinical setting. This study was conducted on a relatively limited dataset of 53 subjects due to the difficulty of obtaining such pediatric imaging data – while finding control subjects is not exceedingly difficult, cases of children with ASD who have had lung CT scans conducted is rather rare. The small sample size was the primary reason that LOOCV was used as the model validation approach. Furthermore, measurements were extracted from these scans through a consistent software process, and the reported results are from models trained on this uniformly calculated data. However, in a clinical setting, measurements may be performed by different practitioners with systematic differences in their techniques, all of which the model may not respond to appropriately. If such a test is to be used in a clinical setting, it must be robust to a wider range of diverse measurement sources, and those performing the measurements must be trained through a consistent process. The clinical interpretability of the machine learning method used here is also doubtful, as there is no direct clinical parallel for the significance of performing PCA projection and SVM classification. However, what can be said is that the subsets of angles determined here may provide some sort of useful information that may serve to indicate ASD.

While the results of this current study suggest a detectable difference in airway geometry between patients with ASD and controls, the need for objective anatomic or biochemical biomarkers to assist with screening for ASD is still clearly unmet. Further investigation is warranted to evaluate airway geometry in ASD, likely with larger sample sizes and categorization of the subtype of ASD given the heterogeneity of the disease. The discrepancies between automated and manual airway angle measurements may also indicate that further work is needed before any clinical considerations can be made. These discrepancies may be due to the subjective nature of when a branchpoint occurs in the airway, as the mathematical model and software used did not take triplicate branching patterns into account.

More powerful machine learning methods and approaches may lead to even better predictive models in future work. Algorithms such as random forest and deep neural networks encompass a much larger hypothesis class and can learn more complex and nonlinear decision boundaries which may yield better classification results. Furthermore, deep learning can be utilized using the original images directly as input rather than angle measurements, as deep neural networks are often able to implicitly perform feature selection and pick up on the most indicative details in the input data.

The differences between the airway of neurotypic and autistic patients were found in airways of the third to fourth generation. These generations are laid down at the end of the embryonic stage, which is shortly before organogenesis of the lung is completed (Schittny, 2017). At this stage, genetic and epigenetic factors lead to the formation of the observed branching patterns including branching points and angles. However, mechanical forces are also important starting early in lung development. Spontaneous contraction of the future airways and fetal breathing movements generate these forces (Koos, 2014; Schittny, 2000), which move fluid through the lung and have an influence on the branching pattern of all airways. Currently little is known about the nervous control of these movements, but we cannot exclude the possibility that neurological differences between neurotypic and autistic children have an influence on these forces and therefore on lung development.

**Conclusion**

There is likely a detectable difference in airway branching angles between children with autism spectrum disorder (ASD) and healthy control subjects. However, this difference's significance and clinical utility is uncertain at this time. There was no detectable doublet morphology of third generation airways on CT scans of the chest. Further investigation is needed to evaluate airway anomalies in children with ASD and determine feasibility as a biomarker for screening and detection of disease.

**Tables**

Table 1: Subject characteristics

|  | Control group | ASD group |
|---|---|---|
| Total number | 23 | 31 |
| Female | 11 | 10 |
| Male | 12 | 21 |
| Age range (years) | 9-20 | 4-30 |
| Average age | 16.2 | 15.0 |
| Standard deviation | 3.3 | 7.4 |

Table 2: Best feature subset of angles by size

| Size | Angles | Accuracy | Sensitivity | Specificity |
|---|---|---|---|---|
| 2 | B1112A1, B1121A2 | 77.36% | 88.57% | 55.56% |
| 3 | B1112A1, B1121A2, B1211A1 | 81.13% | 91.43% | 61.11% |
| 4 | B1112A1, B1121A2, B1211A1, B1212A1 | 86.79% | 94.29% | 72.22% |

| 5 | B1111A1, B1121A1, B112A2, B1121A2, B1122A2 | 86.79% | 97.14% | 66.67% |
| --- | --- | --- | --- | --- |
| 6 | B112A1, B1111A1, B1112A1, B1121A1, B1111A2, B1112A2 | 86.79% | 85.71% | 88.89% |
| 7 | B111A1, B112A1, B1111A1, B1112A1, B1121A1, B1111A2, B1112A2 | 86.79% | 85.71% | 88.89% |
| 8 | B112A1, B1111A1, B111A2, B112A2, B1111A2, B1121A2, B1122A2, B1212A2 | 88.68% | 94.29% | 77.78% |

**Figures**

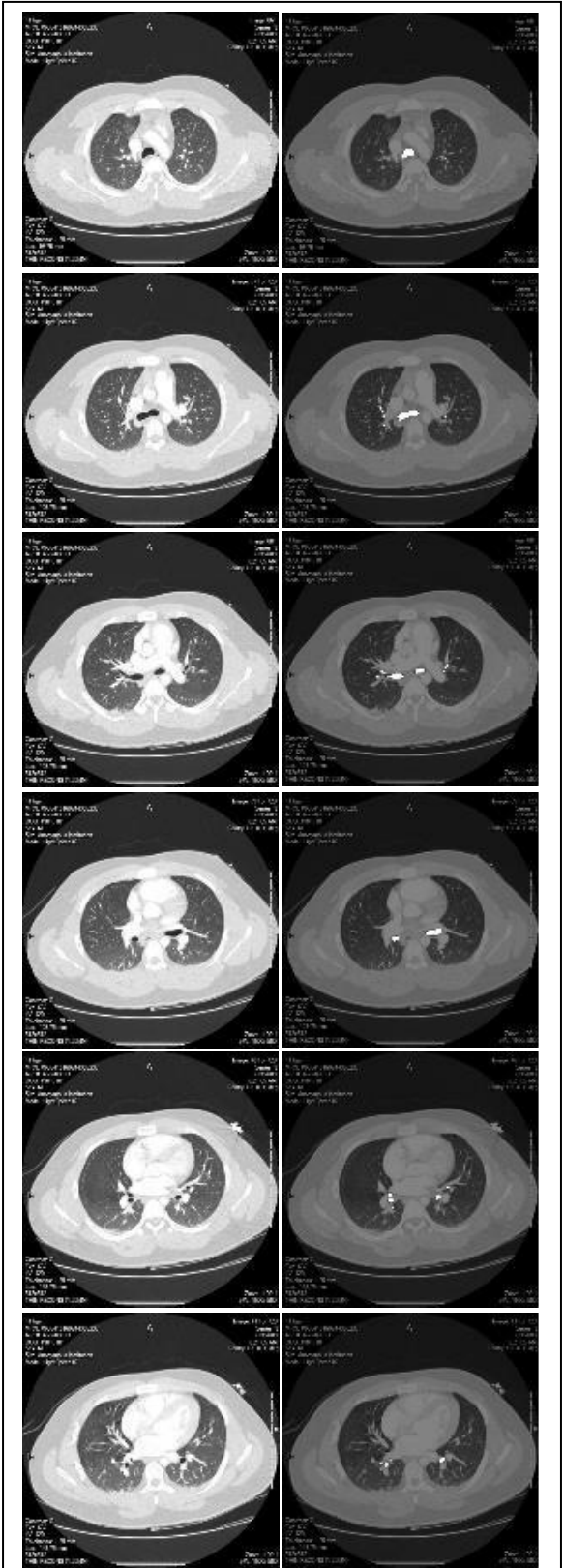

**Figure 1.** Selected slices through a typical chest CT. Left: raw scans. Right: CT scans with airways that have been flood filled with white.

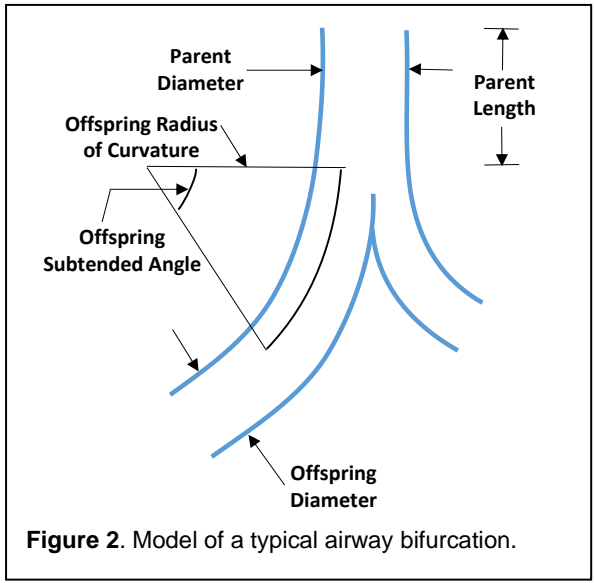

**Figure 2**. Model of a typical airway bifurcation.

```
Begin with 26 branching angles

For s = 1:8

    For each possible subset of size s of the 26 angles

        For k = 1 to s:

            Use PCA to project to k dimensions

            Perform LOOCV using linear SVM

            Save LOOCV Accuracy

        Save k with maximum LOOCV Accuracy
```

**Figure 3.** Pseudocode representation of entire subset sampling, PCA and SVM pipeline.

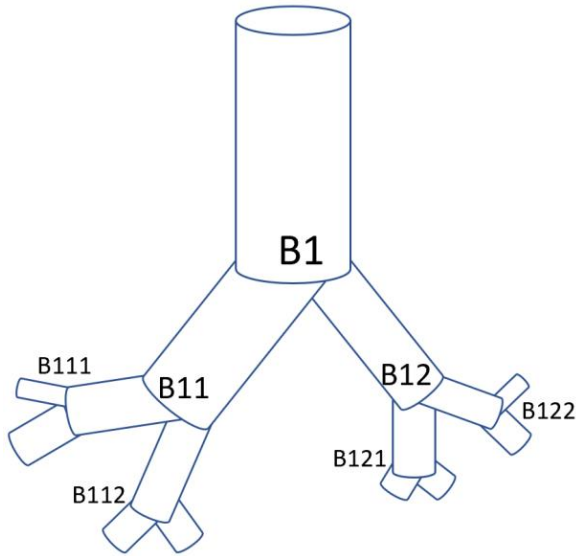

**Figure 4.** Example of labeling sequence of bifurcations. The bifurcation of the trachea into the two bronchi is B1. Thereafter, the label of each daughter branch's bifurcation is created by taking the label of the parent and appending a "1" for the major (larger diameter) daughter and a "2" for the minor (smaller diameter) daughter.

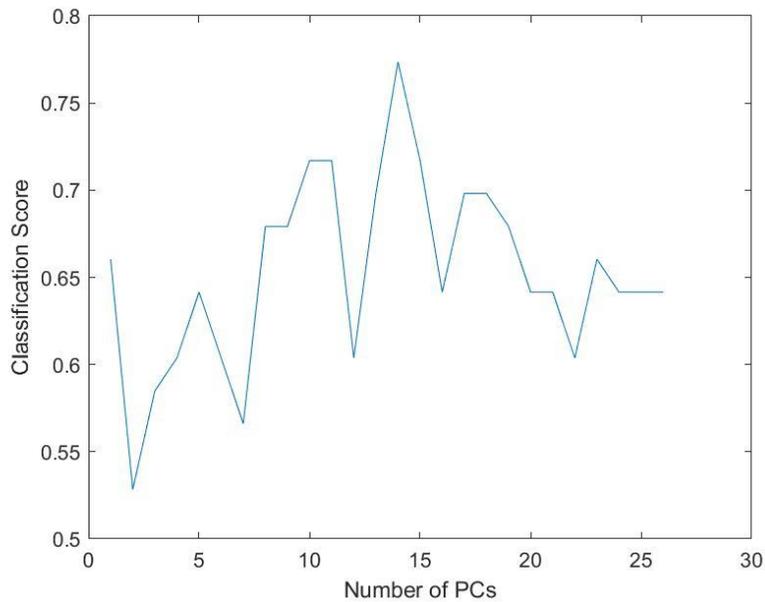

**Figure 5.** Number of Principal Components used vs. LOOCV Classification Accuracy for PCA + Linear SVM based classification procedure using 26 branching angles in first 4 airway generations. In this case, peak accuracy occurs when the number of principal components used is 14.

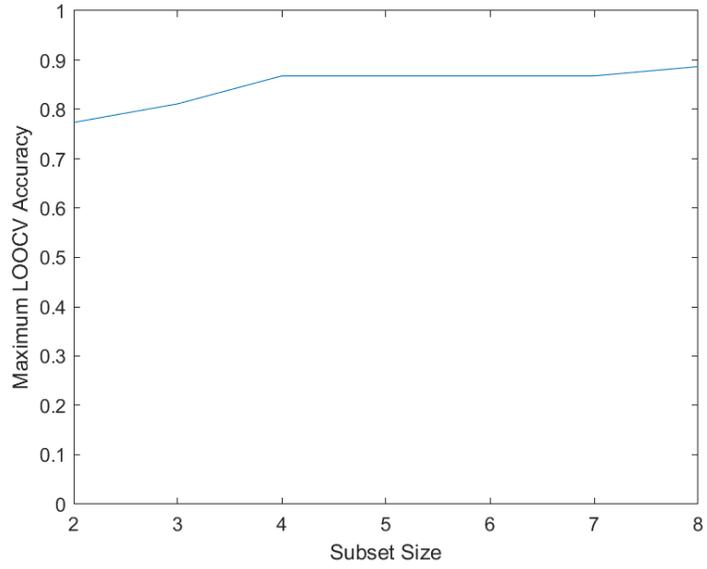

**Figure 6.** Maximum LOOCV accuracy as a function of subset size.